\documentclass[letterpaper, 10 pt, conference]{ieeeconf}  % Comment this line out if you need a4paper

\usepackage{graphicx}
\usepackage{amsmath}
\usepackage{amssymb}
\usepackage{tikz}
\usepackage{pgfplots}
\pgfplotsset{compat=newest}
\usepackage{epigraph}
\usepackage{url}

\IEEEoverridecommandlockouts                              % This command is only needed if 
\title{
Low-Cost Scene Modeling using a Density Function Improves Segmentation Performance
}

\author{Vivek Sharma$^{\diamond \star}$, \c{S}ule Yildirim-Yayilgan$^{\star}$, and Luc Van Gool$^{\diamond \mp}$% <-this % stops a space
\thanks{$^{\diamond}$ ESAT-PSI, VISICS, iMinds, KU Leuven, Kasteelpark Arenberg 10 - box 2441, 3001, Leuven, Belgium,
        {\tt\small \{vivek.sharma, luc.vangool\}@esat.kuleuven.be}}%
\thanks{$^{ \mp}$  Computer Vision Laboratory, ETH Z\"{u}rich, Sternwartstrasse 7, ETH Zentrum, CH - 8092 Z\"{u}rich, Switzerland
        {\tt\small vangool@vision.ee.ethz.ch}}%
\thanks{$^{\star}$ Faculty of Computer Science and Media Technology, Norwegian University of Science and Technology in Gj{\o}vik, Post Box. 191, 2802 Gj{\o}vik, Norway.  
        {\tt\small sule.yildirim@ntnu.no}}%
}

\begin{document}

\maketitle
\thispagestyle{empty}
\pagestyle{empty}

\begin{abstract}
We propose a low cost and effective way to combine a free simulation software
and free CAD models for modeling human-object interaction in order to improve human \& object
segmentation. It is intended for research scenarios related to safe human-robot collaboration (SHRC) and interaction (SHRI) in the industrial domain. The task of human and object modeling has been used for  detecting activity, and for inferring and predicting actions, different from those works, we do human and object modeling in order to learn interactions in RGB-D data for improving segmentation. For this purpose, we define a novel density function to model a three dimensional (3D) scene in a virtual environment (VREP). This density function takes into account various possible configurations of human-object and object-object relationships and interactions governed by their affordances. Using this function, we synthesize a large, realistic and highly varied synthetic RGB-D dataset that we use for training. We train a random forest classifier, and the pixelwise predictions obtained is integrated as a unary  term in a pairwise conditional random fields (CRF). Our evaluation shows that modeling these interactions improves segmentation performance by  $\sim$7\% in mean average precision and recall over state-of-the-art methods that ignore these interactions in real-world data. Our approach is computationally efficient, robust and can run real-time on consumer hardware.
\end{abstract}

\section{Introduction}

Nowadays, in automation industry humans and robots share a common workspace while collaborating on tasks in real-time simultaneously. In such workspaces, any risk of injuries to humans due to the human-robot collaboration should be fully eliminated. For that reason, the robots should have a clear holistic understanding of the scene including the most accurate and precise information related to the human activities in the workspace. This is achievable only if the robotic-system learns the possible context of human-object and object-object  arrangements, and the interactions among them. We therefore  propose a density function that captures the relationship between human-object and object-object interactions. This function can be effectively used to train methods to achieve improved segmentation performance on real-world data. Hence, showing its potential for research scenarios related to safe human-robot collaboration (SHRC) and interaction (SHRI) in the industrial domain.  Optimized modeling for scene understanding workflow eases to learn such contexts.  Objects in real-world interact with each other, so it is very meaningful to capture this interaction for accurate and precise predictions of actions. This is  only possible if the training algorithm (e.g. classifier)  has learnt well  the underlying physical interaction and reasoning for better predictions.  In order to analyse and estimate the scene space in real-world, it is essential to model a synthetic scene relevant to the real-world in a virtual environment for analysis and dataset creation. 

\begin{figure}[t]
\centering
{%\includegraphics[width=0.45\columnwidth]{overview.pdf} \quad
\includegraphics[width=0.58\columnwidth]{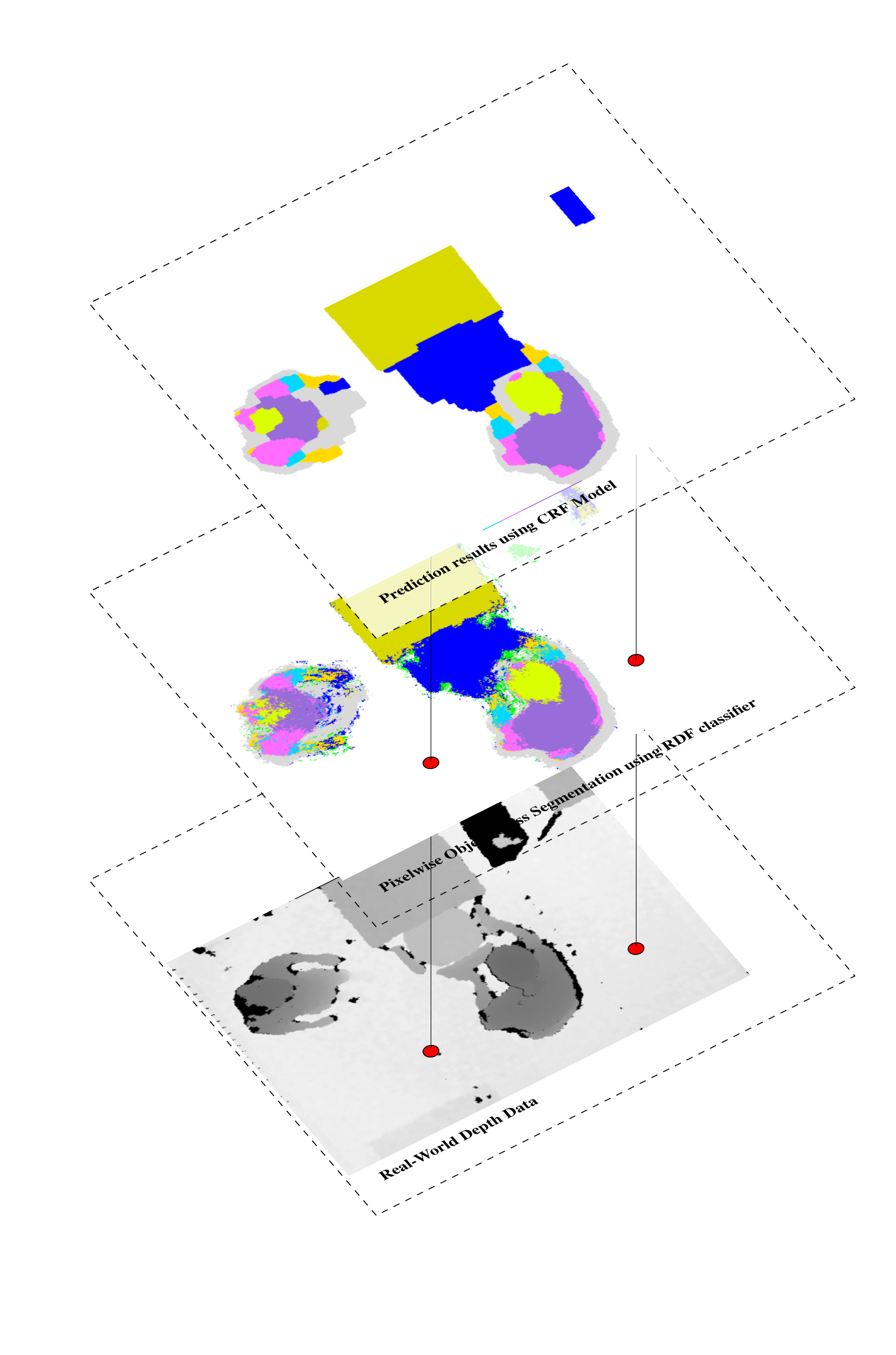}
} \vspace*{-0.15cm}
\caption{Top-view segmentation results for real-world depth data, obtained from KINECT sensor installed at a height of 3.5 meters. (\textit{From bottom:}) The first level of the hierarchy  shows the test depth map, the
second and third levels show the predictions obtained
from RDF classifier and the CRF modeling.} \label{fig:overview_snippet_i}
\end{figure}

In contrast to the past, now we have relatively cheap RGB-D cameras, publicly available free simulation softwares, free 3D CAD models,
low priced consumer hardware with computational resources to allow graphics support. 
In this paper, we keep the cost of hardware and software to barely minimum and focus on achieving an appropriate and efficient real-time segmentation.
We use rendering simulation software (VREP~\cite{vrep}) which is freely available. We use a skeletal tracking system which is based on a multi-sensor KINECT setup instead of using a very expensive marker based motion capture system.  There is minimal manual intervention for generating very large and realistic synthetic scenes by using the rendering software.  Also by using the rendering software we can generate fully automatic ground-truth labeling for free.

In this paper, our goal is to learn the interaction between objects while recognizing objects. This way there is a possibility to infer the scene and predict the actions based on the object modeling. This modeling information is beneficial\footnote{Fraunhofer Institute for Factory Operation and Automation IFF. \url{http://www.iff.fraunhofer.de/en/business-units/robotic-systems/research/human-robot-interaction.html}, 2015}: in aviation/automobile industry for integrating aircraft/automobile components, in manufacturing operations carried hand-in-hand with humans,  for reliable collision detection,  and in healthcare and medical industry for facilitating minimally-invasive-surgery.

For demonstration, we install a RGB-D sensor on the ceiling at the height of 3.5 meters from the ground (see Fig.~\ref{fig:overview_snippet_i}). The camera observes our robotic workspace and the viewed scene is examined for analysis using depth data.  The depth measurements are directly processed to provide an accurate and spatially resolved information about the object classes in real-time from top-view. Our main contributions are as follows:
\begin{itemize}
\item While most previous works consider modeling human and object interactions to infer and predict actions. We propose  to learn  these interactions in RGB-D data in order to improve segmentation which can be useful for SHRI and SHRC scenarios.
\item We propose a  low cost solution composed by a publicly available free rendering simulation software, cheap RGB-D cameras, and free CAD models taken from the internet. With these resources we can create  synthetic scenes that can be used for training and help improve segmentation performance.
\item We show that synthetic data generated from a density function, that governs the human-object and object-object interactions, can be effectively used to train methods to achieve improved segmentation performance by ${\sim}$7\% in mAP and mAR over state-of-the-art methods on real-world data.
\end{itemize}

The remainder of the paper is structured as follows.  In Section II, the related work is given. Section III, describes the proposed density function for scene modeling\iffalse creation of synthetic data\fi. Section IV, explains the feature selection, and some basics on RDF classifier \& CRF modeling. In Section V-VI, implementation details, data collection and the experimental evaluation are given. Finally in Section VII, conclusions are drawn.

\section{Related Work}
For vision applications, we can acquire RGB-D data using real-world sensors or by generating synthetic data using computer graphics rendering techniques in a 3D environment. The real-world data is also possible to use for training the system, but researchers generally avoid this because of several reasons: real-world data generation usually requires expensive equipments (i.e. software and hardware gadgets). In addition manual hand labeling of objects is a very-time consuming and tedious task because precise measurement of objects is required. Also manual ground truth labeling leads to missing out several important information in the scene due to imperfect and error-prone human annotations. Therefore, some vision researchers avoid using real-world data for training vision systems. While by using the present precise rendering techniques and simulation frameworks with modeled environment, it is possible to synthesize accurate and realistic data, which is very close to real-world data. This provides the flexibility to tune the elements of scene for incorporating camera noise, lighting-effects, shadows, rotation/translation, etc. during rendering in order to make the synthetic data as consistent as possible with the physical real-world  data.  

Here, we discuss some of the previous works related to vision problems. In~\cite{Neumann}, Neumann-Cosel~et al. use synthetic data obtained from the modeled Virtual Test Drive simulation for lane tracking in driver assistance and active safety systems. An identical work from Haltakov~et al. ~\cite{Haltakov} proposes a framework for generation of synthetic data from a realistic driving simulator to create  modeled traffic scenarios  similar to real-world scenarios.  A number of works uses synthetic human data for pose and activity recognition~\cite{Dittrich,Me,Ganapathi,Jiang,cvpr,Shotton,Sung}. In vision problems, human pose and activity recognition has received great attention in the last decade. In~\cite{Shotton}, Shotton~et al. synthesize a large and varied dataset of human actions using a motion capture technique.  In~\cite{Me,cvpr}, the authors use a simple synthetic human body representation in a virtual environment using a KINECT skeleton estimation for generating synthetic human pose data. In~\cite{Shotton}, the data is generated based on front, top, and side view, while in~\cite{Me,cvpr}, it is top view only. Both works are applied in real-world scenarios, but neither of the models incorporate the physical interaction of objects with other objects. Also no structural and modeled arrangements are dealt with. Thus, their algorithm fails to recognize  human body-parts, when human body-parts overlap other objects, or are partially occluded by other objects in the scene. In~\cite{Hoiem},  Hoiem~et al. discard the object candidates that are not well supported by proper occlusion boundaries. Yet, we know that reasoning about occlusion would definitely improve the recognition performance. A substantial little work has been done in the context of scene modeling.

Our approach is driven by three key objectives namely computational efficiency, robustness and time efficiency (i.e. real-time) for industrial applications. It further differs from Shotton~et al.~\cite{Shotton} in the following aspects. (a) In~\cite{Shotton}, all training data were thereby synthetically generated by using marker based motion capture (Mo-Cap) technique. Such setups are very expensive and inaccessible to most users. On the other hand, we use a highly optimized virtual representation of 3D human skeleton in a virtual environment. We generate the synthetic data of human body-parts in a virtual environment using a multi-sensor KINECT setup for skeleton tracking in real-world~\cite{Me}. In addition to human data, we also generate synthetic data for objects as well. (b) Calibration of Mo-Cap is a tedious task in comparison to KINECT setup. (c) Processing Mo-Cap data requires sophisticated hardware in comparison to KINECT depth data which can be porcessed with consumer hardware. This way computational expense is reduced.

For learning and reasoning about the physical scene,  we need to capture the relevant and meaningful information (i.e. geometry, pose, shape, occlusion, and other attributes) and train the classifier system. The density functions capture  the relationship, physical interactions, and the context among objects as well as the geometry of the objects.  For physical scene understanding, occlusion boundaries should be well identified, since they form a very crucial component for depth ordering. Incorporating occlusion during synthetic scene modeling, helps to get a better understanding of the real-world scene.  In~\cite{Jiang}, Jiang~et al. model a density function  based on 3D spatial features to capture the object-object and human-object relationship  in a 3D scene. They use the density function to infer arrangements of object in a scene. In~\cite{Jiang}, a camera is installed in the virtual environment, in the corner of the room, where two walls and the ceiling meet, or where a wall and the ceiling meet. Their evaluation is only limited to the virtual environment for training and testing assessments. Similar is done in~\cite{Dittrich}, where Dittrich~et al. model the density function on depth features using a top-view RGB-D sensor. Their evaluation is also limited to their virtual environment for assessment.  Our work is inspired by top-view segmentation from Shotton~et al.~\cite{Shotton} and Sharma~et al.~\cite{cvpr}, and we define the density  function in VREP~\cite{vrep} in order to model object-object and human-object relationships. We use a top-view RGB-D sensor data. However, our evaluation is not just limited to virtual environment, but also to physical real-world scenes. In addition, our work focuses on modeling relationships between a group of multiple 3D objects in a scene, while Shotton~et al.~\cite{Shotton} and Sharma~et al.~\cite{cvpr} only focuses on a single 3D object in a scene. Moreover, they do not incorporate any form of interaction modeling (i.e. non-modeled) between humans and objects.

Some other work, similar to our work, Gupta et al.~\cite{gupta}, Aksoy et al.~\cite{aksoy}, and Pieropan et al.~\cite{pier} use the spatial relationship to perform activities recognition where human-object and object-object relationships are encoded. Also some other notable work, similar to ours that uses the same idea of output of a classifier as input of another model are like ``{Autocontext}" by Tu~et al.~\cite{tu}, ``{Decision Tree Fields}" by Nowotzin~et al.~\cite{now} or ``{Structured Class-Labels in Random Forests for Semantic Image Labelling}" by Kontschieder~et al.~\cite{kont}, among others.

One major problem that arises when using synthetic data, is how to determine if the synthesized data is realistic in comparison to real-world data. This could be determined by evaluating the statistics of object constellation in the synthetic and real world.  A simple approach could be to apply collision avoidance while modeling interactions between objects, but such a modeling is not realistic as occlusions never exist in the modeled scene. Therefore, a realistic scene with appropriate object interactions could be modeled using a density function in order to produce good prediction results, not just for synthetic data, but also for real-world data (see Section~\ref{sec:dataset} and Section~\ref{subsec:comp} for details).

\section{Dataset Generation using a Density Function}\label{sec:dataset}

\begin{figure}[t]
\centering
{%\includegraphics[width=0.45\columnwidth]{overview.pdf} \quad
\includegraphics[width=0.85\columnwidth]{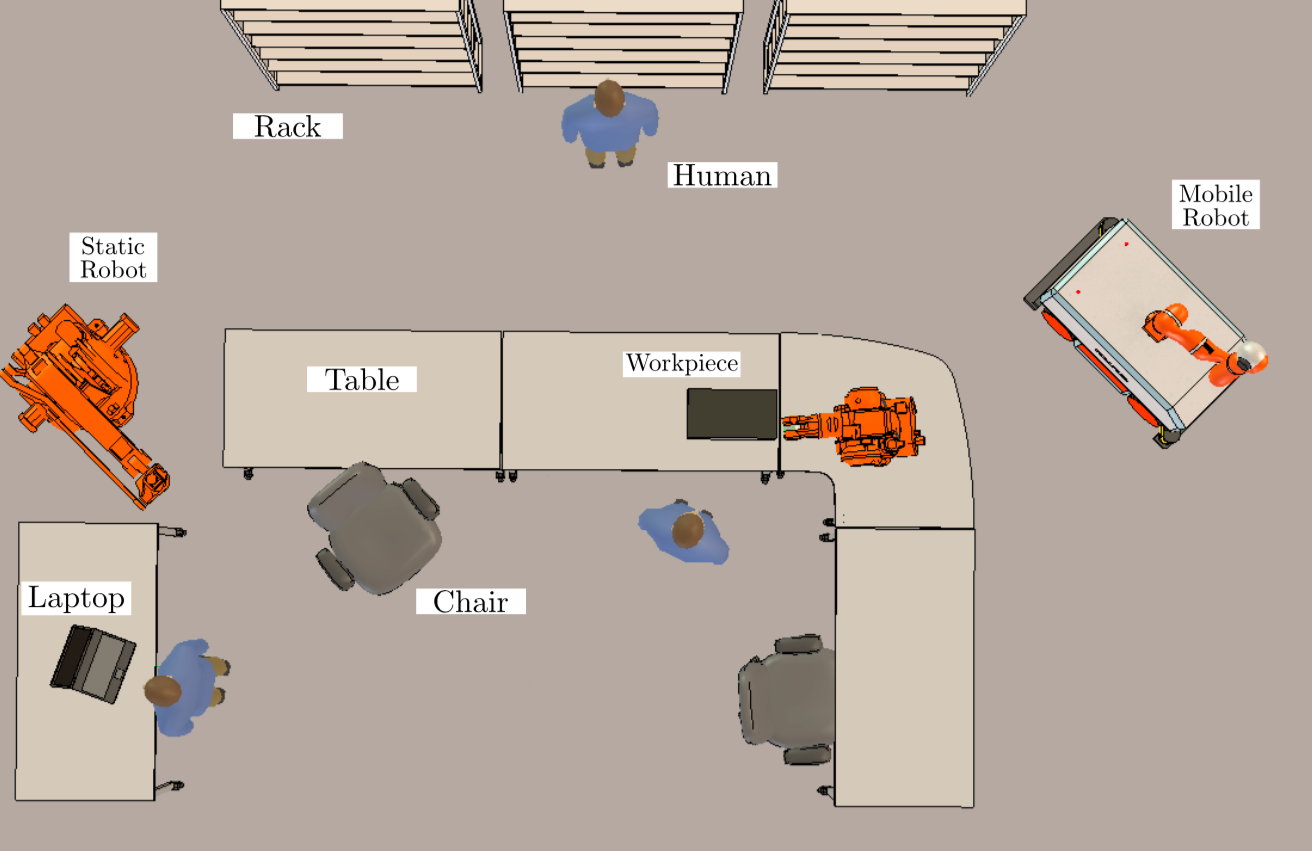}
} \vspace*{-0.15cm}
\caption{Demonstration of a virtual scene in VREP. Object classes from industrial domain is considered as the subset of object constellations  which are consistent to the real-world (Source~\cite{Dittrich})} \label{fig:vrep}
\end{figure}

\begin{figure}[t]
\centering
{\includegraphics[width=0.82\columnwidth]{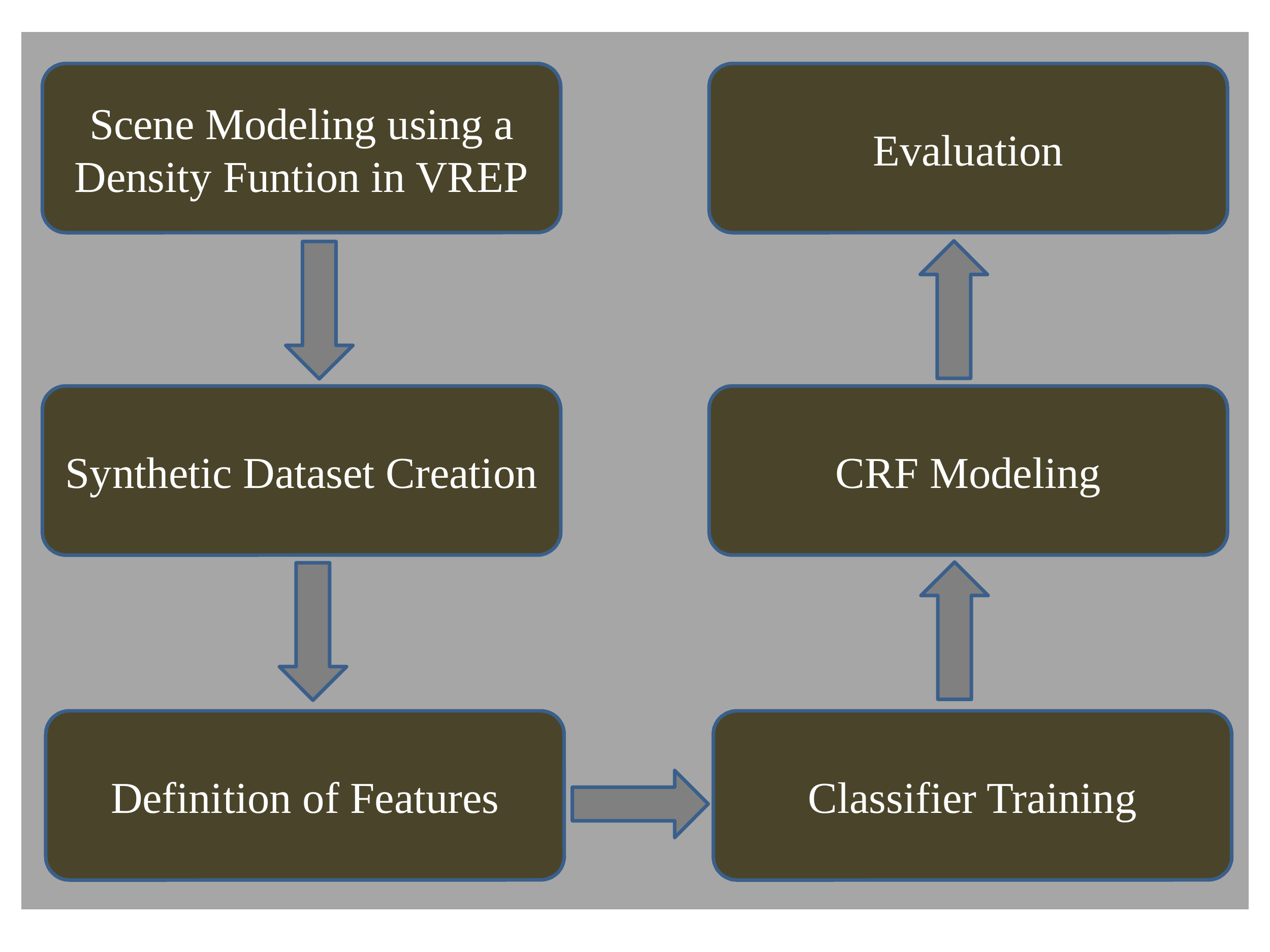} 
} \vspace*{-0.15cm}
\caption{Schematic layout of our proposed segmentation system incorporating: scene modeling using density function in a virtual environment for synthetic data generation; define feature space and correspondingly feature selection and extraction for RDF classifier training; finally CRF modeling.} \label{fig:overview_snippet}
\end{figure}

We used a Virtual Robot Experimentation Platform (VREP) \cite{vrep} (Fig.~\ref{fig:vrep} and Fig.~\ref{fig:overview_snippet}) for modeling and dataset generation. VREP is a robotics simulator with an embedded application for a robot in an integrated development environment (IDE) with potential field tasks without depending physically on the actual robot.  VREP robot simulator has been used for 3D modeling and generation of the synthetic data. VREP simulator provides a virtual environment, which is capable of emulating realistic motion generation, translation, rotation of the actual objects in the IDE. Each object/model can be individually controlled via a remote API client, this makes it ambidextrous because of its distributed control architecture. It supports  C/C++, Python, Java, Lua, Matlab, Octave or Urbi. Also it is publicly available for hobbyists, academic and research purposes. We use 3D CAD models for object classes: chair, table, storage obtained from website\footnote{\url{http://www.hermanmiller.com/design-resources/3d-models-revit/3d-models-by-product/seating.html}, 2015}, and plant (domain: industrial office type) obtained from website\footnote{\url{http://xfrog.com/category/samples.html}, 2015}. These CAD models are also publicly available for free. We created our own 3D human model, and imported it to the virtual environment. The 3D models we use are of the file types *.3ds and *.obj. Refer to \cite{vrep} for detailed information about how to import CAD models into the scene.

\begin{figure*}[htb]
\centering
{\includegraphics[width=1.79\columnwidth]{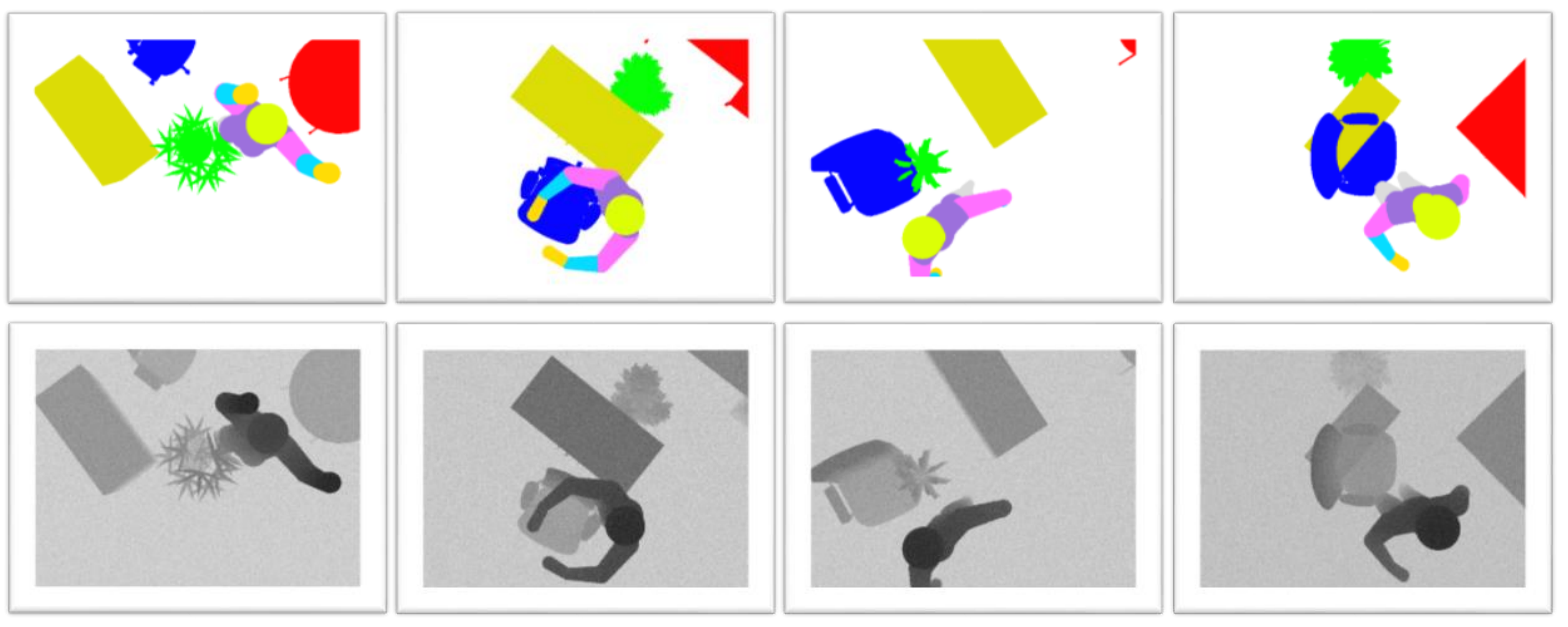}
} \vspace*{-0.15cm}
\caption{ Training dataset.  (\textit{Top}) Ground truth labels  of synthetic depth data (\textit{Bottom}) generated using a density function with a synthetic KINECT sensor. The captured dataset shows high variability of the spatial layout of object, object pose, object orientation, object arrangement, object interaction and relationships between object classes.  Synthetic depth frames are with additive white Gaussian noise.} \label{fig:dataset}
\end{figure*}

Here, we define a density function in VREP for scene modeling, which includes: spatial layout of object, object pose, object orientation, object arrangement, object interaction and relationships between object classes i.e. none, partial, and full occlusion. The density function captures the relationships between  human-object ($H-O$) and object-object ($O-O$) arrangements and the interactions among them. This allows us to generate  a consistent synthetic dataset with physical real-world scenarios. Fig.~\ref{fig:dataset} shows our synthesized dataset using the density function.

A scene is denoted by $S$, and it contains industrial objects $O = \{O_{1},O_{2},...,O_{n}\}$ placed in the scene. The density function captures the relationship of human and industrial objects in the scene that is similar to real-world. This helps  to improve the segmentation performance in  real-world scenarios. 
The human object class is denoted by $H$ with variable poses, specified by spatial joint location and activity.  The density function $\Phi$ capturing the context of human-object and object-object relationships in a scene is defined as: 
\begin{equation}  \Phi(S)=\Psi(H,O;\theta)\Psi(O,O;\theta), \end{equation}
where $\theta$ is threshold. We chose 4 industrial objects (i.e. chair, plant, table, and storage) based on an industrial environment, and 6 localized human body-parts of the human as object classes (i.e. head, body, upper-arm, lower-arm, hand and legs). In order to illustrate the scene well,  the density function describing the human-object and object-object relationships is defined as:
\begin{equation}
\begin{split}
 \Psi(H,O;\theta)= \psi(H_{h})\psi(H_{p})\psi(H_{pos})\psi(H_{ori})\psi(O_{h})\\\psi(O_{pos})\psi(O_{ori})\psi((H,O)_{\theta})\psi((H,O)_{rel})
\end{split}
 \label{eq:h-o}
\end{equation}
\begin{equation} 
\begin{split}
 \Psi(O,O;\theta)= \psi(O_{h})\psi(O_{pos})\psi(O_{ori})\psi((O,O)_{\theta})\\\psi((O,O)_{rel})
\end{split} \label{eq:o-o}\end{equation}
The terms used in the equation~\ref{eq:h-o} and~\ref{eq:o-o} are explained below.

\paragraph*{\textbf{Height ($h$)}} $\psi(H_{h})$ is the candidate parameter for human height, as not all humans are of same body proportions i.e. height and breadth. We set a scaling factor for variable human heights $\{h_{1},h_{2},h_{3},...,h_{n}\}$, that maintains a uniform correlation between height and breadth
of 3D human model with the real-world human. Our 3D human model is modeled on a set of 173 spheres arranged according to the skeleton estimate. We record the real-world human choreographies via KINECT skeleton tracking from a calibrated multi-sensor setup. We use the stored skeletons for modeling the  3D human model in the virtual environment.  The simple scaling of recorded skeleton is a vector  based on body proportions applied as:
\begin{equation}  SC_{scaled} = \beta \cdot SC_{original} = \{\beta \cdot SC_{1}, \beta \cdot SC_{2},...,\beta \cdot SC_{m} \} \end{equation}
where $SC$ is the estimated skeleton setups, $\beta$ being a fixed scaling factor ranging between [$\beta_{min}$ , $\beta_{max}$ ], which maintains a huge variation in body proportions of the 3D human model. This measure facilitates a proper correspondence in the generation of the synthetic human data in approximate relevance to the real-world human. The human height ranges between 160-190 cm. $\psi(O_{h})$ is the candidate parameter for industrial objects height. For table, chair, plant, and storage, standard industrial based heights of object instances were chosen. Table, chair, storage range is between 70-90 cm, and plant ranges between 15-35 cm. The probability density function for   $\psi(H_{h})$  is uniformly distributed on the interval $h \in [h_{1}=160,h_{2}=190]$ and  $\psi(O_{h})$ is uniformly distributed on the interval $h \in [h_{1}=70,h_{2}=90]$.
\iffalse
and is given by:
\begin{equation} \psi(H_{h}) =   \frac{1}{h_{2}-h_{1}},  \enspace    h \in [h_{1}=160,h_{2}=190]   \end{equation} 
and  $\psi(O_{h})$ is uniformly distributed on the interval $h \in [h_{1}=70,h_{2}=90]$.
%\begin{equation} \psi(O_{h})  = \begin{cases}  \frac{1}{h_{2}-h_{1}},  \enspace  for \enspace   h \in [h_{1}=70,h_{2}=90] \end{cases}   \end{equation} 
\fi

\paragraph*{\textbf{Pose ($p$)}} $\psi(H_{p})$ is the candidate parameter for human pose. The human poses and appearances include: \textit{sitting, standing, stretching, walking, working, dancing, bending, bowing, swinging, boxing, and tilting} with combinations of single and both arms, angled arms, and other combinations.  The pre-defined set of poses have one-to-one correspondence to a set of candidate scaling parameter.

\paragraph*{\textbf{Position or translation ($pos$)}} $\psi(H_{pos})$ and  $\psi(O_{pos})$ are the candidate parameters for human and object positions in the scene respectively. There is a preference of human and objects to be placed with reference to each other and translated in the whole parent scene uniformly at random. We incorporate maximum randomization as is the case in real-world. The probability density function for   $\psi(H_{pos})$ and $\psi(O_{pos})$ is uniformly distributed on the interval $pos \in [w,h]$  and is given by $1/(w \times h)$, where $w$ resembles width and $h$ resembles  height of the  parent scene.

\paragraph*{\textbf{Orientation or rotation ($ori$)}} $\psi(H_{ori})$ and  $\psi(O_{ori})$ are the candidate parameters for human and object orientations  in the scene respectively. The rotation ranges between $0 - {360}^{^{\circ}}$ about the vertical axis at random. The probability density functions for  $\psi(H_{ori})$ and $\psi(O_{ori})$ are uniformly distributed on the interval $ori \in [ori_{1}=0,ori_{2}=360]$.

\paragraph*{\textbf{Threshold or distance function ($\theta$)}}  $\psi((H,O)_{\theta})$ and $\psi((O,O)_{\theta})$ are the candidate parameters for defining the threshold of preferred occlusion of boundaries for $H-O$ and $O-O$ classes respectively. This parameter encodes the distance upto what level the $H-O$ and $O-O$ classes would overlap based on the Euclidean thresholded distance ($\theta^{'}$). The probability density functions for   $\psi(H_{h})$ and $\psi(O_{h})$ are uniformly distributed on the interval $\theta \in [\theta_{1}=0,\theta_{2}=\theta^{'}] $.  $\theta^{'}$ can be set to different values for $H-O$ and $O-O$ interactions, though in our case  we use $\theta^{'}=30$ \% for both cases. See ``Comparison of Models'' in Section~\ref{subsec:demon} for more details. 

\paragraph*{\textbf{Relationship ($rel$)}} $\psi((H,O)_{rel})$ and $\psi((O,O)_{rel})$ are the terms that model the relationships for $H-O$ and $O-O$ classes. Relationships like: collision between objects (e.g. 2 tables attached together), objects occluded by one another (e.g. a storage placed below a table), objects  partially occluded (e.g. partial occlusion of a chair by table), objects placed over one-another (e.g. a plant pot placed on top of a table).  Other interactions such as  a person may have his hand placed on the table, holding the chair handle,  sitting on a chair and so on.

\section{Algorithm}

Fig.~\ref{fig:overview_snippet} shows the schematic layout of our proposed hierarchical segmentation system. The synthesized  RGB-D  training dataset incorporates modeled $H-O$ and  $O-O$ relationships and interactions obtained using a density function based scene modeling. The first step performed is sampling i.e. number of frames and features-per-object-class are chosen at random for training an RDF classifier.  Next, individual extracted  features  corresponding to an object class are passed to the RDF classifier. RDF returns a trained classification forest in which each leaf node represents the class prediction of a tree. Now a test depth map of the scene obtained from the real-world KINECT sensor is given as an input to the trained classification forest. The result obtained is a pixelwise object class labeling. The likelihood of an object label assigned to a pixel obtained from the classifier is integrated as a unary term in the pairwise CRF. The pairwise term is obtained from the Potts model~\cite{Potts}. Thus, this labeling problem modeled on pairwise CRF is optimized using $\alpha$-expansion built on graph cuts~\cite{GraphCuts} for finding a global optimal labeling (i.e. segmentation). 

\subsection{Additive white gaussian noise in synthetic depth data}
The synthetic depth data from the virtual KINECT sensor in the scene usually contains no noise. The capturing of the real-world with sensors usually is a combination of ideal signal and noise. The generation of noise may be because of vast number of sources, variations in the detector sensitivity, environmental variations, the discrete nature of radiation, transmission or quantization errors, geometry dependant missing data due to shadows in the IR image cast by the object, that there are discretization errors, and noise increases with increasing depth,   all of which add up to a noise model.  Because of the noise in the real-world
data, and to cope with unseen data samples in the testing step
more robustly, meaning to increase the generalization ability of
the trained classifier, we add additive Gaussian white noise to
the depth values (Fig.~\ref{fig:dataset}).

\subsection{Feature Selection}
We adopt the same depth features as specified in~\cite{Me}. For a given pixel location $s$ of an object sample $O$ from its depth map, we denote its depth value as a mapping $d_{O}(s)$,  and design a feature $f_{O}(s)$ by using two 2D offsets positions $u$, $v$ from $s$:
\begin{equation}  f_{O}(s)= d_{O}(s+u)-d_{O}(s+v)\end{equation}
The  feature is  depth-invariant. We use a rectangular patch for extraction of depth values from an object sample. We compute a fixed number of 300 features for each object sample.

\subsection{Classifier}

The choice of selecting a good discriminative classifier is independent of the preceding sections, because our goal is to show that modeling using density function helps to improve the real-world segmentation performance.  Here,  we use a Random Decision Forests (RDF) classifier for pixelwise object classification. RDF is an ensemble of $t$ binary decision trees, $t \in \{1,...,T\}$. In a decision tree, each of the nodes is associated with a feature response function (i.e. weak learner or split function). The feature response function plays the most critical and crucial role in both training and testing of random forests. In the internal node, selection of parameters of the split function takes place associated with each split node by optimizing a chosen objective function defined on training data set. The objective function is based on maximizing the information gain. The geometric primitives of the split function are used to partition the data points. We employ two geometric primitives, axis-aligned and  linear feature response function. We also use bagging and randomized node optimization, that injects randomness into the trees during the tree training. The stopping criteria used for tree growth were maximum number of levels that a tree could reach, minimum information gain that the nodes could have, child nodes having the same entropy value.

\iffalse
The feature vector is injected at the root node of the decision tree and the first step is randomly choosing a subset of training samples for each training tree (i.e. bagging) in the forest, and randomly proposing a fixed sized subset of feature space dimensions out of $h \times w$ dimensions, i.e. candidate features ($\phi$) and thresholds ($\tau$) per feature for decision in the split node.  The feature vector is split into two subsets, left and right child nodes for each $(\phi, \tau)$ based on the chosen weak learner for the decision tree. Information gain for each ($\phi, \tau$) is computed. The largest information gain $argmax$ for ($\phi, \tau$) is selected and then the feature vector is split. These steps are repeated recursively for all newly constructed child nodes until a stopping criterion for the tree growth is met.  Refer to \cite{me} for detailed explanation. Fig.\ref{fig:RDF} shows a simple pixelwise object class labeling using RDF classifier.
\fi

\subsection{Conditional Random Fields}

The energy of the pairwise CRFs used for object class segmentation  can be defined as a sum of unary  and pairwise  potential terms, given as:
\begin{equation} E(\mathbf{x}) = \sum_{i \in \upsilon} \varphi_{i}(p_{i}) + \sum_{i \in \upsilon, j \in \eta} \varphi_{i,j}(p_{i},p_{j}), \label{eq:CRF_pair_en} \end{equation}  where $\upsilon$ corresponds to the vertex set of a 2D grid with each vertex corresponding  to pixel $p$ in the image and $\eta$ neighborhood of the pixels $\forall  i, j  \in \upsilon$. $\mathbf{x}$ is an arbitrary labeling. The unary  potential $\varphi_{i}(p_{i})$ term in the CRF energy is a data term. It is the likelihood of the object label assigned to pixel $i$. Here, the unary term is obtained from the RDF classifier for each pixel belonging to the object class.  The pairwise potential $\varphi_{i,j}(p_{i},p_{j})$ term in the CRF energy encodes a smoothness prior and encourages smooth segmentation by favoring neighboring pixels in a 2D grid having the same label~\cite{GraphCuts}. It takes the form of a Potts model \cite{Potts}, which is efficiently minimized by $\alpha$-expansion~\cite{GraphCuts} built on graph cuts.

The predictions obtained from pixelwise labeling using RDF is very efficient and when modeled on a CRF to minimize the misclassification of labels assigned to the depth measurements, makes the algorithm more ``robust" and even contributes to the good results.

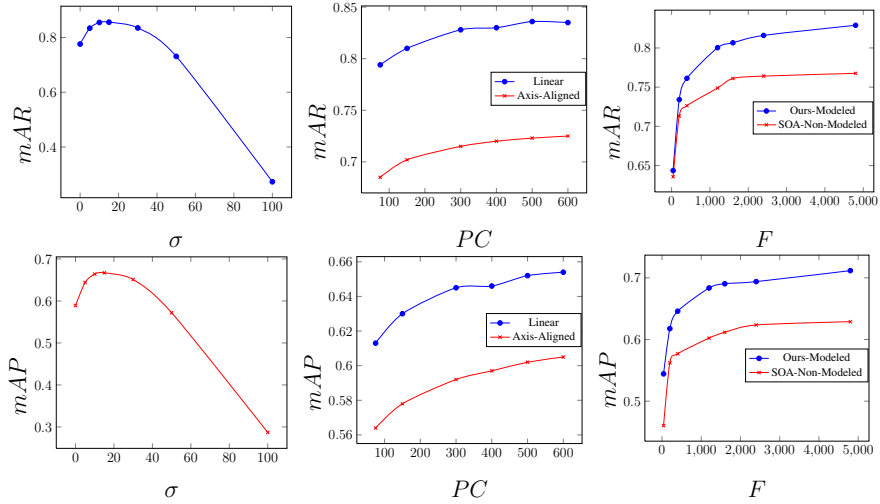
\begin{figure*}[htb]
\centering % centering table

\pgfplotsset{
every axis label/.append style={font=\huge\sffamily},
tick label style={font=\sffamily\large},
}

\resizebox{3.8cm}{!}{
\begin{tikzpicture}
    \begin{axis}[
%      width=0.8\textwidth,
%     height=0.6\textwidth,
	x label style={below=5mm},
   	 y label style={anchor=south east},
        xlabel={$\sigma$},
        ylabel=$mAR$]
    \addplot[smooth,mark=*,blue] plot coordinates {
        (0,0.776)
        (5,0.834)
        (10,0.855)
	(15,0.856)
	(30,0.835)
	(50,0.731)
	(100,0.273)
    };
    \end{axis}
    \end{tikzpicture}}
\resizebox{3.8cm}{!}{
\begin{tikzpicture}
    \begin{axis}[
	x label style={below=5mm},
   	y label style={anchor=south east},
        xlabel={$PC$},
        ylabel=$mAR$, 
	legend style={shift={(0.1,-0.25)}}]
    \addplot[smooth,mark=*,blue] plot coordinates {
	(75,0.794)
	(150,0.81)
	(300,0.828)
	(400,0.83)
	(500,0.836)
	(600,0.835)
    };
    \addlegendentry{Linear}

    \addplot[smooth,color=red,mark=x]
        plot coordinates {
	(75,0.685)
	(150,0.702)
	(300,0.715)
	(400,0.720)
	(500,0.723)
	(600,0.725)
        };
    \addlegendentry{Axis-Aligned}
    \end{axis}
    \end{tikzpicture}}
\resizebox{3.8cm}{!}{
\begin{tikzpicture}
    \begin{axis}[
	x label style={below=5mm},
   	y label style={anchor=south east},
        xlabel={$F$},
        ylabel=$mAR$, 
	legend style={shift={(0.1,-0.40)}}]
    \addplot[smooth,mark=*,blue] plot coordinates {
        (40,0.6436)
        (200,0.7341)
        (400,0.7613)  
	(1200,0.8003)
        (1600,0.8066)
        (2400,0.8160)
	(4800,0.8289)
    };
    \addlegendentry{Ours-Modeled}

    \addplot[smooth,color=red,mark=x]
        plot coordinates {
        (40,0.6357)
        (200,0.7131)
        (400,0.7264)  
	(1200,0.7489)
        (1600,0.7611)
        (2400,0.7641)
	(4800,0.7676)
     };
    \addlegendentry{SOA-Non-Modeled}
    \end{axis}
    \end{tikzpicture}}\\
\resizebox{3.8cm}{!}{
\begin{tikzpicture}
    \begin{axis}[
        x label style={below=5mm},
   	y label style={anchor=south east},
        xlabel={$\sigma$},
        ylabel=$mAP$]
    \addplot[smooth,color=red,mark=x]
        plot coordinates {
        (0,0.589)
        (5,0.644)
        (10,0.664)
	(15,0.667)
	(30,0.651)
	(50,0.572)
	(100,0.287)
        };
    \end{axis}
    \end{tikzpicture}}
\resizebox{3.8cm}{!}{
\begin{tikzpicture}
    \begin{axis}[
	x label style={below=5mm},
   	y label style={anchor=south east},
        xlabel={$PC$},
        ylabel=$mAP$, 
	legend style={shift={(0.1,-0.2)}}]
    \addplot[smooth,mark=*,blue] plot coordinates {
	(75,0.613)
	(150,0.630)
	(300,0.645)
	(400,0.646)
	(500,0.652)
	(600,0.654)
    };
    \addlegendentry{Linear}

    \addplot[smooth,color=red,mark=x]
        plot coordinates {
	(75,0.564)
	(150,0.578)
	(300,0.592)
	(400,0.597)
	(500,0.602)
	(600,0.605)
        };
    \addlegendentry{Axis-Aligned}
    \end{axis}
    \end{tikzpicture}}
\resizebox{3.8cm}{!}{
\begin{tikzpicture}
    \begin{axis}[
	x label style={below=5mm},
   	y label style={anchor=south east},
        xlabel={$F$},
        ylabel=$mAP$, 
	legend style={shift={(0.1,-0.40)}}]
    \addplot[smooth,mark=*,blue] plot coordinates {
        (40,0.5443)
        (200,0.6177)
        (400,0.6459)  
	(1200,0.6836)
        (1600,0.6904)
        (2400,0.6939)
	(4800,0.7117)
    };
    \addlegendentry{Ours-Modeled}

    \addplot[smooth,color=red,mark=x]
        plot coordinates {
        (40,0.4603)
        (200,0.5623)
        (400,0.5770)  
	(1200,0.6023)
        (1600,0.6117)
        (2400,0.6236)
	(4800,0.6289)
        };
    \addlegendentry{SOA-Non-Modeled}
    \end{axis}
    \end{tikzpicture}} \vspace*{-0.15cm}
\caption{ Evaluation results. (\textit{Column 1:}) effect of the additive white Gaussian noise ($\sigma$) to the synthetic training depth data in order to exhibit real-world camera noises on the average recall ($mAR$) and precision ($mAP$) measures of pixelwise object class segmentation. (\textit{Column 2:}) comparison of the linear and axis-aligned feature response function, using $mAR$ and $mAP$ as a function of the number of features extracted ($PC$). \iffalse (\textit{Column 3:}) effect of the number of trees ($T$) in a forest on $mAR$ and $mAP$. \fi (\textit{Column 3:}) comparison of the modeled and non-modeled training dataset, using $mAR$ and $mAP$ as a function of number of training synthetic depth frames ($F$). For the evaluation,  a random number of 65 real-world test depth maps were used.}
\label{fig:evaluation}
\end{figure*}

\section{Implementation Details}

For our implementation, we used  a desktop with Intel i7-2600K CPU at 3.40GHZ, 4GB RAM with an operating system installed on a SSD. A fixed rectangular feature patch of size ($w,h$)=($64,64$) is used for the whole training process. Each frame generated from a KINECT sensor is of size $640\times480$. 

For training our RDF classifier, we used the following parameter setup: (a) tree depth (D=19); (b) number of trees in a forest (T=5); (c) number of features extracted from a rectangular patch (PC=300); (d) number of synthetic scenes (i.e. depth frames) used for training (F=1600/tree); (e) 100 thresholds per feature and 100 feature response samples in the node optimization, along with bagging approach.  

\section{Experimental Evaluation}

The goal of this work is to improve the segmentation performance in real-world scenarios, for this reason we use only real-world data, a random number of 65 test depth maps for evaluation. We acquire RGB-D data from a real-world RGB-D sensor, which is placed on the ceiling in the center of our shared workspace. See Fig.~\ref{fig:dataset_noise_evaluation} (\textit{column 1}) for real-world depth maps.

We evaluate the impact on the segmentation performance  for three cases: (Section \ref{subsec:noise_eval}) what amount of noise should be added to synthetic data to make the synthetic data consistent with real-world; (Section \ref{subsec:feat_eval}) effect of the split function of RDF on classification performance; (Section \ref{subsec:comp}) comparison between modeled and non-modeled training dataset (i.e. with and without density function) through training an RDF classifier, when evaluated on real-world.  

In order to evaluate the effect of modeling human \& object interactions, we show two variants of our method: a modeled version that takes into account these interactions and non-modeled version that ignores them. In addition, we compare against state-of-the-art approaches~\cite{cvpr,Shotton}. Note that, all our experiments are conducted on real-world test data (see Section~\ref{subsec:comp} and Section~\ref{subsec:demon}). For the performance measures, we generate mean average of recall (mAR) and precision (mAP) for each single object class and a combined average of all classes.

\subsection{Data Collection} A scene is a single frame where there is a single object or a combination of multiple 3D objects arranged and oriented in a particular configuration based on the density function. The scenes are composed of complex configurations of humans and objects. We generated an extensive synthetic dataset of RGB-D data using the 3D scene models of 4 different rooms based on industrial workspace and office domain with a total of 10 object classes in the virtual environment~\cite{vrep} (see Fig.~\ref{fig:dataset}). In case of training we use solely synthetic depth data, and in case of the testing data for the evaluation step, we use synthetic and real world depth data. The object classes are: human body-parts (head, body, upper-arm, lower-arm, hand, and legs), table, chair, plant, and storage. We generated the synthetic human body-parts data via using a real-world KINECT multi-sensor setup with a 3D multi-color human model in the virtual environment. For more detailed information about human data generation, refer to previous work~\cite{Me,icml}. We synthesized a dataset of 20,000 frames using density function for $H-O$ and $O-O$ relationships. The synthesized scene has a depth map ranging between $0-3.5$ meters. Each scene is assigned  a set of human, chair, plant, storage, and table object class. Each frame is a stationary image, having no kinematic or temporal information. We use the 3D models of objects based on industrial workspace, with 4 instances for each object class: \textit{chair with and without handles; rectangular and round table; flowers and plants within pot; shelves and wardrobes}.  The modeled scenes are commonly seen in the real-world. In Section~\ref{subsec:comp}, we show that training with synthetic modeled data is sufficient for the generalization of the classifier in regard to real-world data.

\subsection{Noise Evaluation} \label{subsec:noise_eval}
Additive white Gaussian noise using a standard deviation ($\sigma$) was added to synthetic training depth data for compensating with noise in real-world, by matching the camera output as close as possible, and have an approximately good segmentation.  The results in Fig.~\ref{fig:evaluation} (\textit{column 1})  implies  that using this setting for generating
synthetic data results in the best performance, when the synthetic depth frames with additive white Gaussian noise using a standard deviation of 15 cm is used.

\subsection{Feature Response Evaluation} \label{subsec:feat_eval}
The results in Fig.~\ref{fig:evaluation} (\textit{column 2}) show that  most of the gains occur for linear feature response and is  $\sim$10\% more than the axis-aligned. The qualitative results obtained from axis-aligned feature response produced overconfident predictions, while for linear feature response it was also more visually pleasing results.

\subsection{Comparison of Models} \label{subsec:comp} In order to encode a consistent modeling of $H-O$ and $O-O$ interactions, we preferred occlusion of boundaries for maintaining as accurate as possible relevance with the real-world. This relationship between objects employ an euclidean thresholded distance $\theta^{'}$. We used a rank based approach for up to what level the occlusion should be allowed, $\theta^{'}$=\{0, 5, 10, 15, 20, 30, 50\} in \%. We found out that as $\theta^{'}$ is increased, the testing predictions increases monotonically  up to $\theta^{'}$=30, and just after that there is a sharp descent in both quantitative and qualitative performance. Therefore, we preferred $\theta^{'}$=30\%  for overlapping of boundaries, and our dataset was synthesized. 

\begin{figure}[htb]
\centering
{\includegraphics[width=0.832\columnwidth]{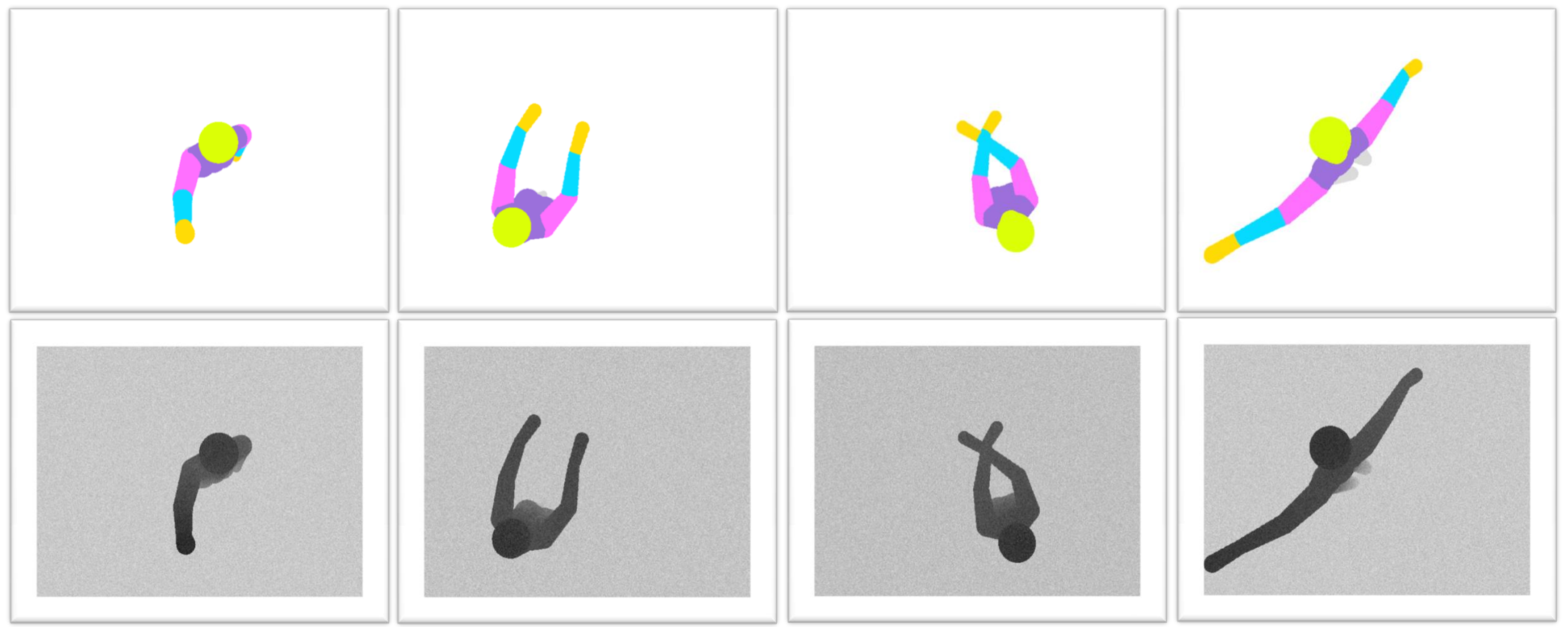}
\includegraphics[width=0.832\columnwidth]{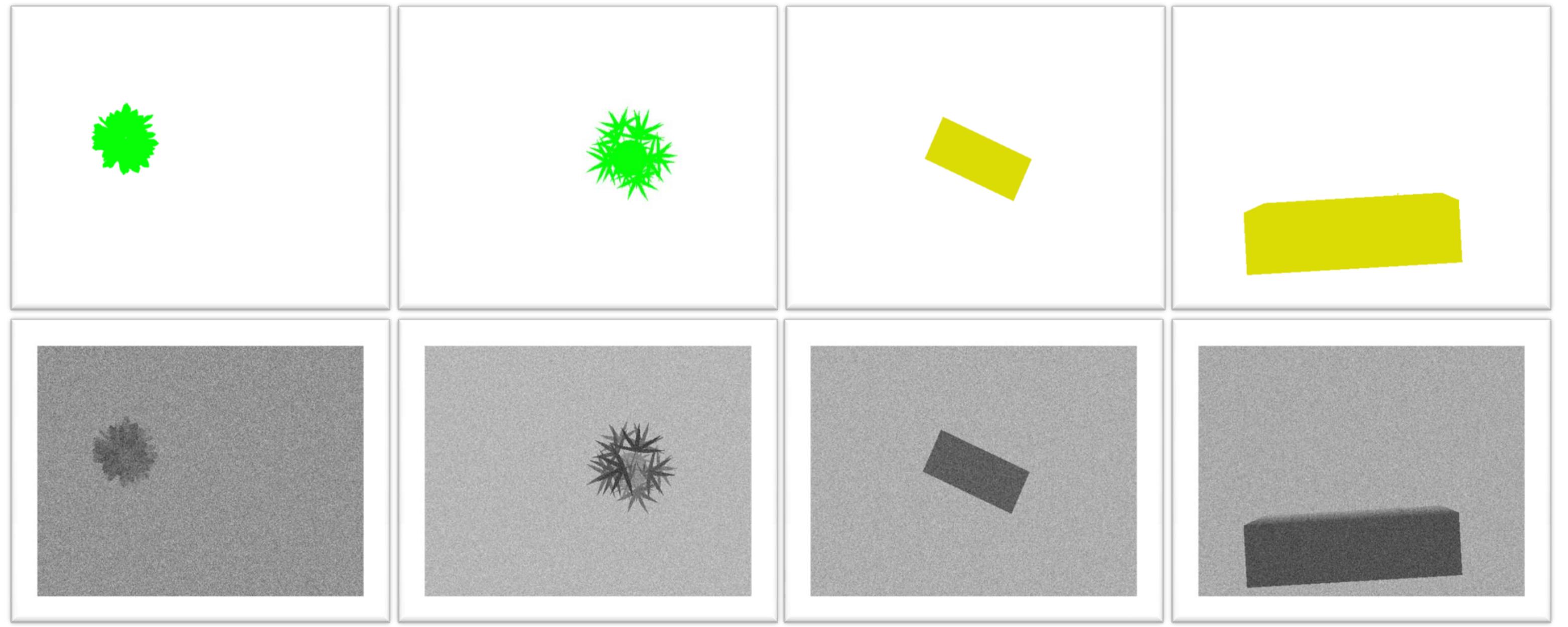}
\includegraphics[width=0.83\columnwidth]{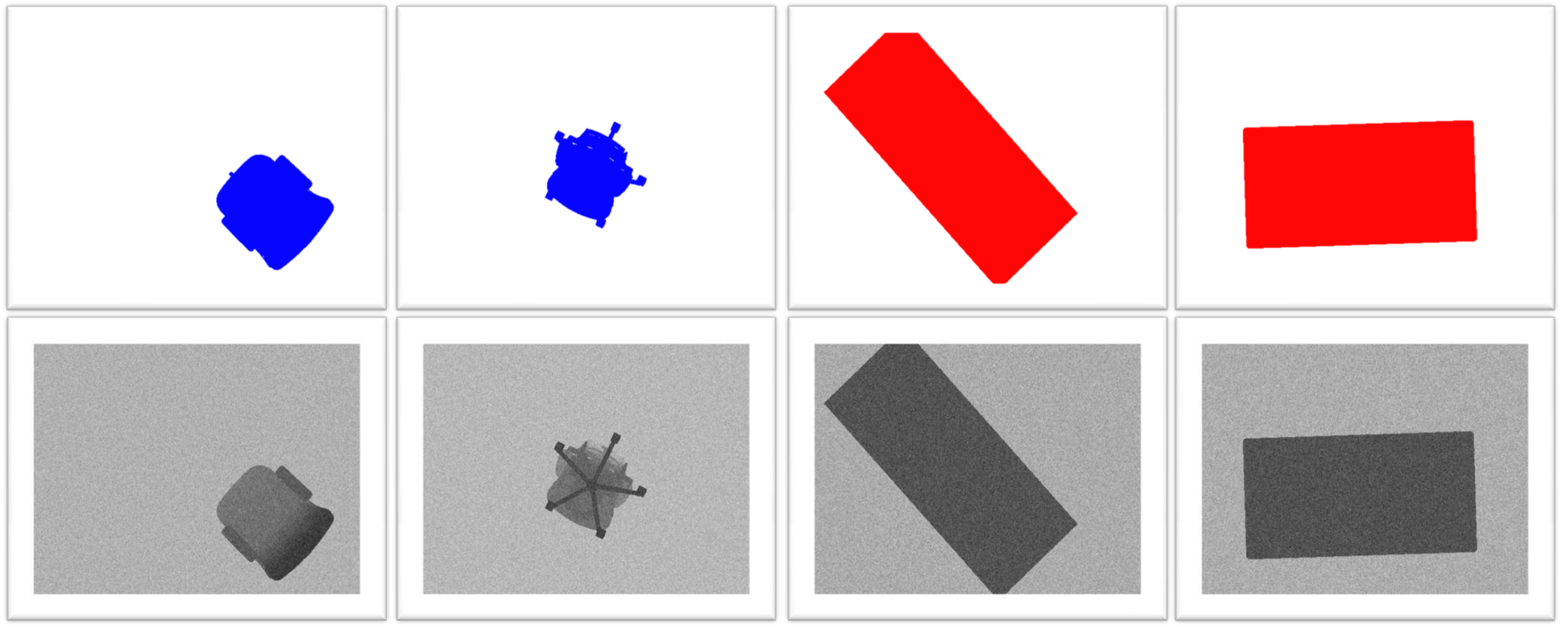}
} \vspace*{-0.15cm}
\caption{ Non modeled training dataset.  Ground truth labels  and its corresponding synthetic depth data generated using the same pipeline used in \cite{Me,Shotton}.} \label{fig:dataset_non}
\end{figure}

\begin{figure}[b]
\centering
{
\includegraphics[width=0.87\columnwidth]{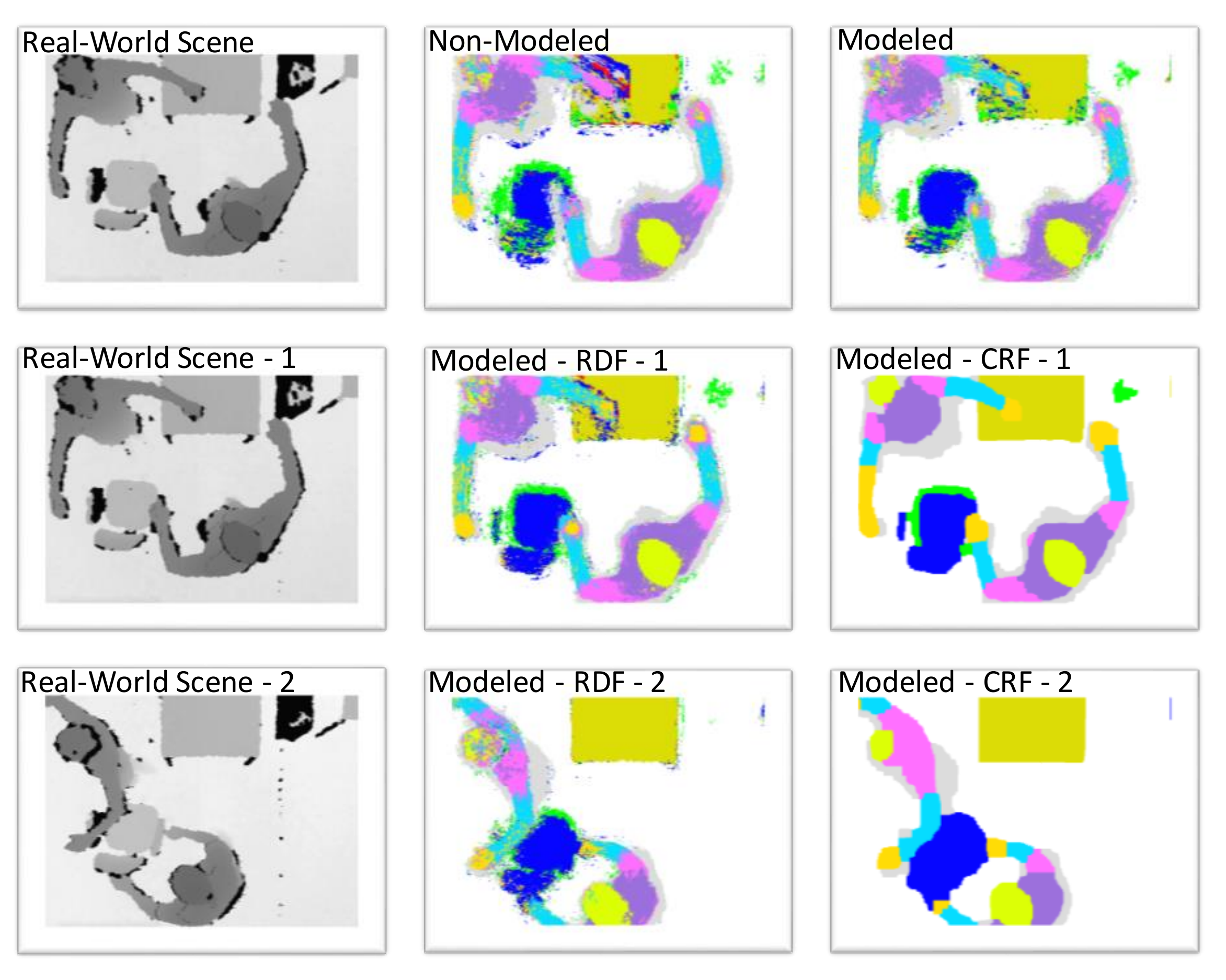}
} \vspace*{-0.15cm}
\caption{ (\textit{Row 1:}) prediction results for real-world test depth data using the modeled and non-modeled training dataset. The segmentation improvements can be seen in the modeled case: the misclassification around the border of the human has diminished significantly; the human hand placed on the table and the chair are classified well with reduced mislabeling. (\textit{Row 2-3:}) shows the predictions obtained from the RDF classifier and the CRF modeling.} \label{fig:dataset_noise_evaluation}
\end{figure}

As a baseline, we implemented the same  state-of-the-art (SOA) pipeline used in \cite{cvpr,Shotton} for the generation of the non-modeled dataset, based on top-view (see Fig.~\ref{fig:dataset_non}). In a non-modeled scene, there is only a single 3D object in a scene with a particular configuration. In our case, there are multiple 3D objects in a scene which are modeled using a density function. We compare the results obtained with our modeled dataset against the results obtained by their non-modeled dataset using a number of frames for training the classifier. The results depicted in Fig.~\ref{fig:evaluation} (\textit{column 3}) thereby show  a steep ascent with a higher $mAR$ and $mAP$  for our proposed modeling. The density function based modeling, substantially improves the performance by  $\sim$7\% in  $mAR$ and $mAP$ over state-of-the-art~\cite{cvpr,Shotton}. Fig.~\ref{fig:dataset_noise_evaluation} (\textit{row 1}) shows the reduced misclassification around the border of the human body with better classification of the human-hand placed on the table and the chair for modeled case.

\begin{table*}[t]
\caption{Confusion matrix-based mean  average F1-measure} \vspace*{-0.15cm}
\centering % centering table
\resizebox{16cm}{!}{
\begin{tabular}{|r|l l l l l l l l l l l|}
  \hline
 F1-measure & Avg  & Head& Body & UArm & LArm & Hand & Legs & Chair & Plant & Storage & Table \\
  \hline 
$CRF_{Non-Modeled}$~\cite{cvpr,Shotton} &{0.76} &0.90 &0.71 &0.73 &0.65 &0.69 &0.48 &0.85 &0.78 &0.90 &0.91 \\
$CRF_{Modeled}$ &\textbf{0.84} &\textbf{0.96} &\textbf{0.84} &\textbf{0.79} &\textbf{0.70} &\textbf{0.79} &\textbf{0.52} &\textbf{0.93} &\textbf{0.90} &\textbf{0.98} &\textbf{0.97}\\
  \hline
\end{tabular}}
 \label{table:optimization_precision}
\end{table*}

\subsection{Demonstration} \label{subsec:demon}
We use the integrated system for segmentation in our shared robotic workspace in real-time using real-world depth data. Fig.~\ref{fig:dataset_noise_evaluation} (\textit{row 2-3}) shows the prediction results for the proposed CRF model using real-world test data.  Table~\ref{table:optimization_precision} shows that the  F1-measure for the CRF  modeled  improves by  $\sim$8\% over the non-modeled ones~\cite{cvpr,Shotton}. For training the model with our optimized RDF parameter setup takes 43 minutes, and testing the model takes 34 ms.  Each block of our pipeline (i.e. scene modeling using a density function, fine tuning the additive noise in the synthetic data, selection of  the linear feature response for the RDF tree training, modeling the labeling problem on a CRF to minimise the misclassification of labels)  contributes  to the good results, and plays a crucial role in improving the segmentation performance. The resulting system is computationally efficient, robust and supports real-time for our targeted application.

\section{Conclusion}

In this paper we described an inexpensive and effective way to model the human-object and object-object interactions in order to improve  segmentation performance. For this purpose, we proposed a density function to model a 3D scene in a virtual environment for synthesizing a dataset incorporating human-object and object-object interactions, that is consistent with the real-world scenarios. Our proposed density function models spatial layout of object, object pose, object orientation, object arrangement, object interaction and relationships between the object classes i.e. none, partial, and full occlusion by each-other. Our experiments are based on industrial workspace and office domain with a total of 10 object classes. Our goal of modeling scene using density function improves the real-world scene segmentation performance measures by  $\sim$7\% in mean average precision and recall over state-of-the-art non-modeling based segmentation methods.  Our integrated resulting  system is computationally efficient, robust and supports real-time for our targeted application. In future work, we would like to extend this work towards  the semantic image understanding and image-sentence generation~\cite{Fei,Farhadi} for manufacturing and automation industry of challenging environments for safe human-robot interaction and collaboration.

\section*{Acknowledgment}

We would like to thank Jos{\'e} {Oramas M}, Michele Fenzi, Frank Dittrich and Stephan Irgenfried for their valuable suggestions.

\normalsize
\bibliographystyle{ieee}
\bibliography{egbib}

\end{document}